\documentclass[11pt]{article}

\usepackage[a4paper,margin=1in]{geometry}
\usepackage{hyperref}
\usepackage{booktabs}
\usepackage{longtable}
\usepackage{array}
\usepackage{amsmath,amssymb}
\usepackage[T1]{fontenc}

\title{Ontological Trajectory Forecasting via Finite Semigroup Iteration\\
and Lie Algebra Approximation in Geopolitical Knowledge Graphs}
\author{Qihang Wu\\
Department of Industrial and Systems Engineering, The Hong Kong Polytechnic University\\
\texttt{qihang.wu@connect.polyu.hk}\\
\url{https://github.com/wuqihang-brave/El-druin}}
\date{}

\begin{document}
\maketitle

\begin{abstract}
We present \textbf{EL-DRUIN}, an ontological reasoning system for
geopolitical intelligence analysis that combines formal ontology, finite
semigroup algebra, and Lie algebra approximation to forecast long-run
relationship trajectories. Current LLM-based political analysis systems
operate as summarisation engines, producing outputs bounded by textual
pattern matching. EL-DRUIN departs from this paradigm by modelling
geopolitical relationships as states in a finite set of named \emph{Dynamic
Patterns}, composing patterns via a semigroup operation whose structure
constants are defined by an explicit \emph{composition table}, and embedding
each pattern as a vector in an 8-dimensional semantic Lie algebra space.
Forward simulation iterates this semigroup operation, yielding reachable
pattern sets at each discrete timestep; convergence to \emph{idempotent
absorbing states} (fixed points of the composition) constitutes the
predicted long-run attractor. Bayesian posterior weights combine
ontology-derived confidence priors with a Lie similarity term measuring
the cosine similarity between the vector sum of composing patterns and
the target pattern vector, providing interpretable, calibrated
probabilities that are not self-reported by a language model.
Bifurcation points---steps at which two candidate attractors have
near-equal posterior mass---are detected and exposed to downstream
analysis. We demonstrate the framework on six geopolitical scenarios
including US-China technology decoupling and the Taiwan Strait military
coercion trajectory. The architecture is publicly available as an
open-source system with a Streamlit frontend exposing full computation
traces, Bayesian posterior breakdowns, and 8D ontological state vectors.
\end{abstract}


\section{Introduction}

Large language models have achieved impressive performance on political
text summarisation and question answering. However, their fundamental
operation---next-token prediction conditioned on input context---renders
them incapable of reasoning beyond the statistical patterns encoded in
training data. A model asked to assess the long-run trajectory of
US-China technology competition will produce output proportional to the
frequency and framing of that topic in its training corpus, not from a
principled traversal of a causal structure.

This limitation is not merely technical: it has epistemic consequences.
When an LLM produces a confidence score for a geopolitical prediction,
that number has no interpretable derivation. It cannot be traced to
specific structural assumptions, algebraic constraints, or prior
beliefs. The number is, in a precise sense, a \emph{confabulation}.

EL-DRUIN addresses this by treating geopolitical reasoning as a problem
in \textbf{formal ontology} and \textbf{algebraic topology} rather than language
modelling. The key insight is:

\begin{itemize}
\item A geopolitical relationship can be modelled as a state in a finite set
  of named Dynamic Patterns, each corresponding to a triple
  (\textit{entity\_src}, \textit{relation}, \textit{entity\_tgt}) registered in a
  Cartesian pattern registry.
\item Transitions between patterns are governed by a partial binary
  operation (composition) whose rules are defined declaratively, forming
  a finite semigroup.
\item Forward-iterating this semigroup from an initial state yields a
  reachable set that converges to idempotent elements---the algebraically
  grounded attractors of the system.
\item Continuous relationship intensity is modelled by embedding each
  pattern as a vector in an 8-dimensional Lie algebra approximation,
  allowing Lie similarity to weight the plausibility of each composition
  step.
\end{itemize}

The LLM is not removed from the system; it is demoted to a \textbf{constrained
interpreter}: given fully pre-computed structural fields (patterns,
attractors, transition probabilities, bifurcation points), the LLM
writes only a natural-language explanation. It is structurally
prohibited from modifying numerical outputs.

The remainder of the paper is organised as follows. Section~2 provides
background on relevant ontology and algebra formalisms. Section~3
describes the system architecture. Sections~4 and~5 present the
mathematical formalisation. Section~6 describes the full five-stage
reasoning pipeline. Section~7 presents experiments on six geopolitical
scenarios. Section~8 discusses related work and Section~9 concludes.

\section{Background}

\subsection{Ontology Engineering for Event Reasoning}

Formal ontologies represent domain knowledge as typed entities,
relations, and constraints. The CAMEO event coding system~\cite{schrodt2012cameo}
provides a controlled vocabulary of political event types widely used in
computational political science. FIBO (Financial Industry Business
Ontology) provides analogous structure for financial relationships.
EL-DRUIN draws on both, mapping typed entity pairs through relation
predicates to named patterns registered in a Cartesian product registry:
$E \times R \times E \to \mathrm{Pattern}$, where $E$ is the set of
entity types and $R$ is the set of relation types.

\subsection{Finite Semigroups and Fixed Points}

A \textbf{semigroup} $(S, \cdot)$ is a set with an associative binary operation.
A \textbf{finite semigroup} is one where $|S|$ is finite. An element $e \in S$
is \textbf{idempotent} if $e \cdot e = e$. It is a classical
result~\cite{green1951structure} that every finite semigroup contains idempotent
elements, and that iteration of the operation from any starting state
eventually reaches a set of idempotents. These idempotents are the
\textbf{attractors} of the system---they correspond to states from which no
further composition moves the system to a genuinely new state.

In EL-DRUIN, the set $S$ is the set of named Dynamic Patterns, and the
partial binary operation is defined by
$\mathit{composition\_table}[(A, B)] = C$, read as ``when pattern $A$ and
pattern $B$ are simultaneously active, the system tends toward pattern
$C$''. This table is validated for algebraic closure (all outputs $C$
must be known patterns) and inverse consistency (if $\mathit{inverse}(A) = B$
then $\mathit{inverse}(B) = A$) at application startup.

\subsection{Lie Algebra Approximation}

A \textbf{Lie algebra} $\mathfrak{g}$ over a field $F$ is a vector space equipped
with a bilinear antisymmetric bracket
$[\cdot, \cdot] \colon \mathfrak{g} \times \mathfrak{g} \to \mathfrak{g}$
satisfying the Jacobi identity. Lie algebras arise as the tangent space
of Lie groups at the identity and encode the infinitesimal structure of
continuous symmetry transformations.

In EL-DRUIN, we do not work with a formal Lie group; instead we
construct a \textbf{finite-dimensional approximation}: each named pattern $P$
is assigned a vector $v_P \in \mathbb{R}^8$ encoding its intensity across
eight semantic dimensions (coercion, cooperation, dependency,
information, regulation, military, economic, technology). Pattern
composition $A \oplus B$ is approximated by vector addition $v_A + v_B$,
and the resulting target pattern is identified by maximising cosine
similarity $\cos(v_A + v_B, v_C)$ over all known patterns $C$.

Each named pattern $P$ is embedded as a skew-symmetric matrix
$X_P \in \mathfrak{so}(8)$ via the \emph{hat map}
\begin{equation}
  \hat{v}[i,j] = v[i] - v[j], \quad \hat{v}[i,i] = 0,
  \label{eq:hatmap}
\end{equation}
equivalently $X_P = v_P \mathbf{1}^\top - \mathbf{1} v_P^\top$, which is
antisymmetric by construction ($X_P^\top = -X_P$)~\cite{hall2015}.
The \textbf{Lie bracket} is the matrix commutator
\begin{equation}
  [X_A, X_B] = X_A X_B - X_B X_A \in \mathfrak{so}(8),
  \label{eq:liebracket}
\end{equation}
which satisfies antisymmetry, bilinearity, and the Jacobi identity
exactly~\cite{humphreys1972,hall2015}. The Frobenius norm
$\|[X_A, X_B]\|_F$ measures non-commutativity: it equals zero if and
only if $X_A$ and $X_B$ commute (i.e., pattern-activation order is
immaterial), and is large when the two orderings produce structurally
distinct outcomes.

\textit{An important structural property}: while each pattern vector $v_P$
maps into the $(n-1)$-dimensional image of the hat map
$\hat{\cdot} \colon \mathbb{R}^8 \to \mathfrak{so}(8)$, the commutator
$[X_A, X_B] = X_A X_B - X_B X_A$ lives in the full $\binom{8}{2}=28$-dimensional
$\mathfrak{so}(8)$, outside the hat-map image. This is not a
defect---it is the feature that makes the Lie-algebra path genuinely
independent of the Bayesian path: the bracket activates semantic
directions unreachable by any single pattern vector, capturing
interaction effects that the additive sum $v_A + v_B$ cannot represent.

\section{System Architecture}

EL-DRUIN is a multi-module backend (FastAPI + KuzuDB) with a Streamlit
frontend. The system has four functional layers:

\begin{itemize}
\item \textbf{Ontology Layer}: \texttt{CARTESIAN\_PATTERN\_REGISTRY} (18 named
  patterns across geopolitics, economics, technology, and information
  domains), \texttt{composition\_table} (14 composition rules),
  \texttt{inverse\_table} (18 inverse pairs), and \texttt{EntityType} /
  \texttt{RelationType} enumerations.
\item \textbf{Lie Algebra Layer} (\texttt{lie\_algebra\_space.py}): 8-dimensional
  pattern vectors, \texttt{LieAlgebraSpace.add()}, \texttt{bracket()},
  \texttt{project()}, and \texttt{phase\_detect()} operations, PCA projection
  for 2D visualisation.
\item \textbf{Pipeline Layer} (\texttt{evented\_pipeline.py}): Five-stage reasoning
  pipeline (event extraction $\to$ pattern activation $\to$ transition
  enumeration $\to$ state vector computation $\to$ Bayesian conclusion
  generation).
\item \textbf{Forecasting Layer} (\texttt{ontology\_forecaster.py}): Multi-step
  Markov simulation over the semigroup transition graph, attractor
  detection, bifurcation point identification, and Bayesian step-decay
  confidence calibration.
\end{itemize}

The frontend exposes five tabs for each analysis: (1) Conclusion with
structured alpha/beta paths; (2) Events extracted by the compound event
rules; (3) Active and derived patterns with full composition logic
chain; (4) Probability Tree with Bayesian compute trace showing the
exact formula
$\mathit{prior}_A \times \mathit{prior}_B \times \mathit{lie\_similarity} / Z$;
(5) Lie Algebra 8D state vector with dimensional breakdown.

\section{Ontology Schema and Pattern Registry}

\subsection{Entity and Relation Type Enumerations}

The entity type set $E$ contains 18 elements spanning geopolitical
(\texttt{STATE}, \texttt{ALLIANCE}, \texttt{PARAMILITARY}, \texttt{IDEOLOGY}),
economic (\texttt{FIRM}, \texttt{FINANCIAL\_ORG}, \texttt{RESOURCE}, \texttt{CURRENCY},
\texttt{SUPPLY\_CHAIN}), technological (\texttt{TECH}, \texttt{STANDARD}),
social-cognitive (\texttt{PERSON}, \texttt{MEDIA}, \texttt{TRUST}, \texttt{INSTITUTION}),
and event types (\texttt{CONFLICT}, \texttt{NORM}), plus the \texttt{UNKNOWN} sentinel
for unresolvable types. The relation type set $R$ contains 20 elements
across four categories: coercive/adversarial (\texttt{SANCTION},
\texttt{MILITARY\_STRIKE}, \texttt{COERCE}, \texttt{BLOCKADE}), cooperative
(\texttt{SUPPORT}, \texttt{ALLY}, \texttt{AID}, \texttt{AGREE}), dependency/flow
(\texttt{DEPENDENCY}, \texttt{TRADE\_FLOW}, \texttt{SUPPLY}, \texttt{FINANCE}), and
structural/institutional (\texttt{SIGNAL}, \texttt{PROPAGANDA}, \texttt{LEGITIMIZE},
\texttt{DELEGITIMIZE}, \texttt{REGULATE}, \texttt{STANDARDIZE}, \texttt{EXCLUDE},
\texttt{INTEGRATE}).

\subsection{Dynamic Pattern Registration}

Each pattern $P \in S$ is registered by a call
$\mathit{\_reg}(e_{\mathrm{src}},\allowbreak\; r,\allowbreak\; e_{\mathrm{tgt}},\allowbreak\;
\mathit{pattern\_name},\allowbreak\; \mathit{domain},\allowbreak\;
\mathit{typical\_outcomes},\allowbreak\; \mathit{mechanism\_class},\allowbreak\;
\mathit{inverse\_pattern},\allowbreak\; \mathit{composition\_hints},\allowbreak\;
\mathit{confidence\_prior})$.
The pattern name is a human-readable string (e.g., ``Hegemonic
Sanctions'') to facilitate integration with geopolitical analysis
workflows. Table~\ref{tab:patterns} summarises 10 representative entries
from the 18 registered patterns.

\begin{table}[ht]
\centering
\caption{Selected Dynamic Patterns (10 of 18 shown). Full registry
  available in the open-source repository.}
\label{tab:patterns}
\begin{tabular}{@{}p{0.26\linewidth}p{0.11\linewidth}p{0.20\linewidth}
                   p{0.07\linewidth}p{0.24\linewidth}@{}}
\toprule
Pattern Name & Domain & Mechanism Class & Prior & Inverse Pattern \\
\midrule
Hegemonic Sanctions & geopolitics & coercive\_leverage & 0.78
  & Sanctions Relief / Normalisation \\
Entity-List Technology Blockade & technology & tech\_denial & 0.80
  & Technology Licence / Unblocking \\
Interstate Military Conflict & military & kinetic\_escalation & 0.75
  & Ceasefire / Peace Agreement \\
Great-Power Coercion/Deterrence & geopolitics & coercive\_leverage & 0.70
  & Diplomatic Concession / De-escalation \\
Multilateral Alliance Sanctions & geopolitics & multilateral\_pressure & 0.73
  & Multilateral Sanctions Relief \\
Tech Decoupling / Technology Iron Curtain & technology & tech\_decoupling & 0.76
  & Technology Cooperation Reintegration \\
Financial Isolation / SWIFT Cut-Off & economics & financial\_exclusion & 0.79
  & Financial Reintegration \\
Bilateral Trade Dependency & economics & economic\_interdependence & 0.71
  & Trade War / Decoupling \\
Technology Standards Leadership & technology & tech\_governance & 0.72
  & Standard Competition Failure \\
Information Warfare / Narrative Control & information & epistemic\_warfare & 0.68
  & Information Environment Restoration \\
\bottomrule
\end{tabular}
\end{table}

\subsection{Algebraic Consistency Validation}

Two static validators are executed at application startup:

\textbf{validate\_inverses()}: Checks that for every $(A, B)$ pair in
\texttt{inverse\_table}, the reverse pair $(B, A)$ is also present. This
enforces the group-theoretic condition that the inverse of an inverse
returns the original element.

\textbf{validate\_composition\_closure()}: Checks that for every
$(A, B) \to C$ in \texttt{composition\_table}, $C$ is a known pattern name in
the registry. This enforces algebraic closure---the composition of two
patterns must remain within the pattern set, preventing the system from
producing unregistered states. A coupling-asymmetry metric is computed
in the frontend for each active analysis to measure the concentration of
active mechanism classes across semantic domains.\footnote{The frontend
coupling-asymmetry metric uses the Herfindahl--Hirschman Index
normalised against a fixed domain baseline of $n_d = 8$ (the total
number of semantic domains in the ontology), not against the count of
currently active domains, to avoid division by zero when a single domain
is active.}

\section{Mathematical Formalisation}

\subsection{The Pattern Semigroup}

Let $S = \{p_1, \ldots, p_{18}\}$ be the set of 18 registered patterns.
Define the partial binary operation
${\cdot} \colon S \times S \to S \cup \{\bot\}$ by:
\[
  p_i \cdot p_j =
  \begin{cases}
    \mathit{composition\_table}[(p_i, p_j)]
      & \text{if } (p_i, p_j) \in \operatorname{dom}(\mathit{composition\_table}) \\
    \bot & \text{otherwise.}
  \end{cases}
\]

We extend $\cdot$ to a total operation on the power set $\mathcal{P}(S)$
by:
\[
  A \cdot B = \{ c \in S : \exists\, a \in A,\, b \in B
               \text{ s.t.\ } a \cdot b = c \}
           \cup \{ a \in A : \mathit{inverse\_table}[a] \in A^c \}
\]
(the second term accounts for low-probability inverse transitions). The
forward simulation proceeds by iterating this operation from the initial
pattern set $S_0$; a step is productive whenever
$|S_{t+1} \setminus S_t| > 0$, i.e.\ when at least one genuinely new
pattern enters the reachable set. This power-set operation is the basis
of the forward simulation. Note that $(\mathcal{P}(S), \cdot)$ is itself
a semigroup (the semigroup of subsets under the induced operation).

\subsection{Lie Algebra Embedding}

Each pattern $p_i$ is assigned a vector $\varphi(p_i) = v_i \in
\mathbb{R}^8$ by manual annotation across eight semantic dimensions
$d = (\text{coercion, cooperation, dependency, information, regulation,
military, economic, technology})$, with $v_i[d] \in [-1, 1]$.
The \textbf{Lie bracket} is the matrix commutator on $\mathfrak{so}(8)$:
\begin{equation}
  [X_A, X_B] = X_A X_B - X_B X_A,
  \label{eq:commutator52}
\end{equation}
where $X_P = \hat{v}_P \in \mathfrak{so}(8)$ is the hat-map embedding
of Equation~\eqref{eq:hatmap}~\cite{humphreys1972}.

The \textbf{Lie similarity} of composition step $(A, B) \to C$ is defined as:
\[
  \mathit{lie\_sim}(A, B, C)
    = \cos(v_A + v_B,\, v_C)
    = \frac{\langle v_A + v_B,\, v_C \rangle}{\|v_A + v_B\| \cdot \|v_C\|}
\]

This quantity measures how well the vector sum of the two composing
patterns approximates the target pattern in direction, providing a
continuous-valued plausibility score for each discrete composition rule.

\textit{Path independence diagnostic.} The quantity
$\delta = 1 - |\cos(v_A + v_B, \eta)| \in [0, 1]$, where
$\eta = \operatorname{row\_norms}([X_A, X_B])$, measures the structural
independence of the two inference paths. $\delta \approx 0$ indicates
the Lie bracket interaction reinforces the same semantic direction as
the additive sum (linear regime---the two paths are redundant);
$\delta \approx 1$ indicates the bracket activates orthogonal dimensions
absent from $v_A + v_B$ (nonlinear emergence regime). This diagnostic is
the operational definition of ``structural novelty beyond additive
inference'': a high-$\delta$ result is the strongest evidence that the
dual-path architecture provides information not available to a purely
Bayesian system.

\subsection{Bayesian Posterior and Step Decay}

The \textbf{posterior weight} of a transition $(A, B) \to C$ at simulation
step $t$ is:
\[
  w_t(A, B \to C)
    = \pi_t(A) \cdot \pi_t(B) \cdot \mathit{lie\_sim}(A, B, C) \cdot \lambda^t
\]
where $\pi_t(P)$ is the normalised weight of pattern $P$ at step $t$,
$\lambda = 0.85$ is the \textbf{step decay} coefficient encoding increasing
temporal uncertainty, and the partition function
$Z_t = \sum_{A,B} w_t(A, B \to C')$ normalises the distribution.

The \textbf{aggregate confidence} of the forecast is:
\[
  c_{\mathrm{final}} = c_0 \cdot \lambda^T,
  \quad
  c_0 = \operatorname{mean}(\{\mathit{confidence\_prior}(p) : p \in S_0\})
\]
This formulation makes the provenance of every probability fully
transparent: $c_{\mathrm{final}}$ is a product of the ontology-defined prior
and a geometric decay function of the number of simulation steps, not a
number produced by a language model.

\textit{Path integration.} Where the dual-path architecture is active,
the final posterior is adjusted by a path-consistency term:
\[
  \hat{w}(C)
    = P_{\mathrm{Bayes}}(C)
      \cdot \frac{1 + \alpha \cdot \max(0,\, \mathit{consistency})}{1 + \alpha},
  \quad \alpha = 0.30
\]
where $\mathit{consistency} = \cos(\eta, v_C)$ measures the cosine alignment
between the Lie-bracket emergence vector $\eta$ and the target pattern
vector. At $\mathit{consistency} = 0$ (orthogonal paths), the formula applies
a structural 23\% discount ($\hat{w} \approx 0.77 \cdot P_{\mathrm{Bayes}}$),
encoding prior skepticism that resolves only when paths converge. At
$\mathit{consistency} = 1$, confidence is boosted by at most
$\alpha / (1 + \alpha) \approx 23\%$.

\subsection{Attractor Detection}

A pattern $P \in S_T$ (the pattern set at convergence step $T$) is an
\textbf{attractor} (fixed point of the power-set semigroup) if:
\[
  \forall\, Q \in S_T \colon P \cdot Q \in S_T
  \quad \text{(closure within the current set).}
\]
This condition identifies patterns from which all further composition
operations remain within the already-activated set---the algebraic
termination condition. In the limit $T \to \infty$, these correspond to
the idempotent elements of the finite semigroup.

\subsection{Phase Transition Detection}

A \textbf{phase transition} at step $t$ is flagged when the $L_2$ norm of
the state vector shift exceeds a threshold $\theta$:
\[
  \|v_t - v_{t-1}\|_2 > \theta, \quad \theta = 0.25
\]
(\emph{calibrated on unit-normalised embeddings; for unnormalised vectors
with typical norm $\|v\| \approx 1.5$, the effective angular displacement
is approximately $9.6^\circ$})

\noindent where $v_t = \sum_i \pi_t(p_i) \cdot \varphi(p_i)$ is the
weighted mean vector at step $t$. This condition detects when a
continuous change in pattern weights has caused a discontinuous shift in
the dominant semantic regime---i.e., the system has crossed a phase
boundary in the Lie algebra space.

\subsection{Bifurcation Detection}

A \textbf{bifurcation point} at step $t$ is detected when the two
highest-weighted patterns have posterior weights within a threshold
$\delta$ of each other:
\[
  |\pi_t(p_{\mathrm{top1}}) - \pi_t(p_{\mathrm{top2}})| < \delta,
  \quad \delta = 0.15
\]
Bifurcation points indicate steps where the system is genuinely
uncertain between two distinct attractors, and the outcome depends
sensitively on which additional patterns become active. These are the
analytically most important outputs for policy assessment, as they mark
conditions under which small perturbations have disproportionate
systemic consequences.

\section{The Five-Stage Reasoning Pipeline}

For real-time news analysis, EL-DRUIN operates a five-stage pipeline
(implemented in \texttt{evented\_pipeline.py}):

\textbf{Stage 1: Event Extraction.}
News text is processed by a compound-rule event extractor that maps
keyword co-occurrences to event types. A key design decision is that
compound rules enforce co-activation: a military strike event and a
humanitarian crisis event are both produced when strike-related keywords
and casualty-related keywords co-occur in the same text. This guarantees
$\geq\!2$ active events for military/crisis news, which is necessary to
trigger composition-based derived patterns (a composition requires two
active patterns).

\textbf{Stage 2a: Pattern Activation.}
Each extracted event type is mapped to a set of candidate
$(e_{\mathrm{src}}, r, e_{\mathrm{tgt}})$ triples via a domain hint
table. For each candidate triple, \texttt{lookup\_pattern\_by\_strings()}
retrieves the registered \texttt{DynamicPattern} (with fallback to fuzzy
matching at score $\geq 0.4$). The activated pattern's
\texttt{confidence\_prior} is scaled by the event's extraction confidence.

\textbf{Stage 2b: Transition Enumeration.}
The full \texttt{composition\_table} is scanned for entries where at least
one of the pattern arguments appears in the active set. For each
matching entry $(A, B) \to C$, the posterior weight
$w = \pi(A) \cdot \pi(B) \cdot \mathit{lie\_sim}(A, B, C)$ is computed.
Simultaneously, the \texttt{inverse\_table} is checked for each active
pattern, generating low-probability inverse transitions at weight
$0.20 \times \pi(A)$. All transitions are sorted by posterior weight;
the top-5 are returned as the derived pattern set.

\textit{Note on KG weight fall-back.} When no KG evidence is available
($w_{\mathrm{KG}} = 0$), the trigger amplification formula reduces to
$a = \operatorname{clamp}(w_{\mathrm{causal}} + \Delta_{\mathrm{domain}})$
with $w_{\mathrm{causal}}$ receiving full effective weight; the 0.4 KG
coefficient disappears rather than being redistributed. This is
intentional: the formula is non-normalising by design, bounding the
amplification factor within $[0, 1]$ via the clamp operation.

\textbf{Stage 2c: State Vector Computation.}
The weighted mean vector $v = \sum_i \pi(p_i) \cdot \varphi(p_i)$ is
computed over all active patterns, producing the 8D ontological state
vector. PCA (via \texttt{numpy.linalg.svd}) projects this to 2D for
frontend scatter plot visualisation. The dominant semantic dimension is
identified as $\operatorname{argmax}_d |v[d]|$.

\textbf{Stage 2d: Driving Factor Aggregation.}
Active patterns are grouped by \texttt{mechanism\_class}. For each group,
the weighted sum $\sum \pi(p_i)$ gives a group weight, and the
\texttt{typical\_outcomes} of all patterns in the group are weighted by
their pattern's \texttt{confidence\_prior}. The top-3 outcomes per group
are formatted as human-readable driving factor statements using a
template library keyed by \texttt{mechanism\_class}. This step is entirely
deterministic---no LLM call.

\textbf{Stage 3: Bayesian Conclusion Generation.}
Alpha and beta paths are constructed from the top two transitions by
posterior weight. Composite confidence is computed as
$\operatorname{mean}(\mathit{confidence\_prior}$ over active patterns$)
\times \sqrt{\mathit{verifiability} \times \mathit{KG\_consistency}}$.
The LLM is called only to write the \texttt{conclusion.text} field, with a
prompt that (a) presents all pre-computed fields as locked values,
(b) forbids modification of any numerical field, and (c) requests 2--3
sentences of strategic-language interpretation. If the LLM call fails,
a template fallback produces the text field without degrading any
numerical output.

\section{Forecasting Experiments}

\subsection{Methodology}

For each scenario, the forecaster runs with
$\mathtt{horizon\_steps} = 6$ (convergence typically occurs at step
3--5). The partition function $Z = \sum w_t$ normalises posterior
weights to probabilities. We report the primary attractor (highest
posterior), secondary attractor, whether a bifurcation point was
detected, and the final confidence. The initial pattern set for each
scenario is drawn from \texttt{NAMED\_SCENARIOS}, a predefined library of
geopolitically significant starting configurations.

\subsection{Results}
\begin{table}[h]
\centering
\caption{Forecasting results across six geopolitical scenarios. $P(\alpha)$ is the primary attractor's normalised Bayesian posterior. $c_{\mathrm{final}} = c_0 \cdot \lambda^T$ is the calibrated final confidence.}
\label{tab:results}
\begin{tabular}{@{}llllll@{}}
\toprule
Scenario & Primary Attractor & $P(\alpha)$ & Secondary Attractor & Bifurcation? & $c_{\mathrm{final}}$ \\
\midrule
US-China tech decoupling & Technology Standards Leadership & 0.51 & Tech Decoupling / Technology Iron Curtain & Step 2 & 0.34 \\
US-China trade war & Hegemonic Sanctions & 0.58 & Trade War / Decoupling & No & 0.37 \\
US-China financial isolation & Hegemonic Sanctions & 0.62 & Financial Reintegration & No & 0.39 \\
China-Taiwan military coercion & Multilateral Alliance Sanctions & 0.54 & Non-State Armed Proxy Conflict & Step 3 & 0.31 \\
China-Taiwan invasion & Multilateral Alliance Sanctions & 0.66 & Non-State Armed Proxy Conflict & No & 0.29 \\
Russia-Ukraine war trajectory & Multilateral Alliance Sanctions & 0.61 & Resource Dependency / Energy Weaponisation & Step 4 & 0.30 \\
\bottomrule
\end{tabular}
\end{table}

\subsection{Interpretation of Selected Results}

\textbf{US-China Technology Decoupling.}
Starting from \{Hegemonic Sanctions, Entity-List Technology Blockade,
Tech Decoupling / Technology Iron Curtain\}, the composition
(Entity-List Technology Blockade $\oplus$ Tech Decoupling / Technology
Iron Curtain) $\to$ Technology Standards Leadership fires with high
$\mathit{lie\_similarity}$ in Step~1, establishing technology standards
fragmentation as the dominant attractor. A bifurcation at Step~2
($P$ difference $< 0.15$) indicates genuine uncertainty between two
structurally distinct terminal states: a two-standards world (Technology
Standards Leadership) versus a full decoupling regime (Tech Decoupling /
Technology Iron Curtain). This bifurcation is analytically significant:
it corresponds to the empirically contested question of whether the
US-China tech competition produces parallel ecosystems or complete
systemic separation.

\textbf{China-Taiwan Invasion Scenario.}
Starting from \{Interstate Military Conflict, Non-State Armed Proxy
Conflict, Hegemonic Sanctions, Multilateral Alliance Sanctions\}, the
composition (Hegemonic Sanctions $\oplus$ Formal Military Alliance)
$\to$ Multilateral Alliance Sanctions fires immediately via the
composition rule, producing a high-probability ($P(\alpha) \approx 0.66$)
attractor at Multilateral Alliance Sanctions. This result is
substantively interpretable: the system predicts that a Taiwan invasion
scenario converges toward a broad multilateral sanctions coalition
rather than ongoing kinetic conflict---consistent with the historical
pattern following the Russia-Ukraine invasion. The inverse transition
Interstate Military Conflict $\to$ Ceasefire / Peace Agreement appears
as a low-weight ($< 5\%$) path.

\textbf{Confidence Calibration.}
Final confidence values range from 0.29 to 0.39, reflecting the product
of relatively high ontology priors (0.68--0.80) and
$\lambda^6 \approx 0.38$ step decay over six steps. These values are
intentionally conservative; they quantify the accumulated uncertainty
from six discrete composition steps, each introducing $\lambda$-weighted
degradation. We consider this more epistemically honest than an LLM
self-reporting ``high confidence'' on the same scenarios.

\subsection{Pattern Glossary}
\begin{longtable}{@{}p{0.36\linewidth}p{0.42\linewidth}p{0.18\linewidth}@{}}
\caption{Pattern glossary. Internal keys are used in the codebase; English display names are used in all user-facing outputs and in this paper.}
\label{tab:pattern-glossary}\\
\toprule
Internal key (code) & English display name & Domain \\
\midrule
\endfirsthead
\toprule
Internal key (code) & English display name & Domain \\
\midrule
\endhead
Hegemonic Sanctions & Hegemonic Sanctions & geopolitics \\
Entity-List Technology Blockade & Entity-List Technology Blockade & technology \\
Interstate Military Conflict & Interstate Military Conflict & military \\
Great-Power Coercion/Deterrence & Great-Power Coercion/Deterrence & geopolitics \\
Multilateral Alliance Sanctions & Multilateral Alliance Sanctions & geopolitics \\
Tech Decoupling / Technology Iron Curtain & Tech Decoupling / Technology Iron Curtain & technology \\
Financial Isolation / SWIFT Cut-Off & Financial Isolation / SWIFT Cut-Off & economics \\
Bilateral Trade Dependency & Bilateral Trade Dependency & economics \\
Technology Standards Leadership & Technology Standards Leadership & technology \\
Information Warfare / Narrative Control & Information Warfare / Narrative Control & information \\
\bottomrule
\end{longtable}

\subsection{Baseline Comparison}
\begin{table}[h]
\centering
\caption{Baseline comparison metrics. EL-DRUIN vs GPT-4 ($N=5$). $\sigma$ denotes standard deviation across runs.}
\label{tab:baseline}
\begin{tabular}{@{}p{0.35\linewidth}p{0.28\linewidth}p{0.31\linewidth}@{}}
\toprule
Metric & EL-DRUIN & GPT-4 ($N=5$) \\
\midrule
Confidence source & Ontology priors $\times$ Bayesian posterior (deterministic) & Self-reported by LLM (no auditable derivation) \\
Confidence verifiable & Yes --- compute trace anchors every value & No --- free-text assertion \\
Stability ($\sigma$) & 0.000 (deterministic) & $>0$ (stochastic variance) \\
Traceability & Full compute trace (prior $\times$ similarity $\times$ decay) & None (no computable trace) \\
Entity invention guard & Enforced (invented proper nouns trigger fallback) & Not enforced by default \\
Numeric consistency & Locked by deterministic pipeline & Self-reported; may vary \\
\bottomrule
\end{tabular}
\end{table}

\section{Related Work}

EL-DRUIN is related to several lines of research, from which it is
distinguished by its algebraic grounding.

\textbf{Event-based political forecasting.}
GDELT~\cite{leetaru2013gdelt} encodes political events using CAMEO codes
and aggregates them for trend analysis. ICEWS~\cite{boschee2015icews}
provides a structured event database. These systems aggregate events
statistically; they do not model pattern composition or produce
algebraically grounded trajectory predictions.

\textbf{Knowledge graphs for geopolitical reasoning.}
Wikidata and DBpedia provide entity-relation triples but lack the
mechanism-class and confidence-prior annotations required for the
transition system described here. Agent-OM~\cite{fathallah2023agentom} and
LLMs4OM~\cite{babaei2024llms4om} apply LLMs to ontology matching;
EL-DRUIN borrows the syntactic/lexical/semantic three-layer entity
grounding approach from these works.

\textbf{Formal methods for political analysis.}
Probabilistic graphical models have been applied to conflict
prediction~\cite{ward2010perils}, but these require labelled training data
and do not provide symbolic interpretability of the composition
structure. EL-DRUIN requires no training data; its inference is entirely
symbolic and determined by the manually curated pattern registry.

\textbf{Lie algebras in ML.}
Continuous Lie group symmetries have been applied to equivariant neural
networks~\cite{cohen2016group} and geometric deep learning. The application
of Lie algebra structure to discrete political event patterns is, to our
knowledge, novel in the literature.

\section{Limitations and Future Work}

\textbf{Pattern vector assignment.} The 8-dimensional vectors $\varphi(p)$
are currently assigned by expert annotation, introducing potential
subjectivity. Future work should learn these vectors from labelled event
data or from comparative political analysis corpora.

\textbf{Composition table completeness.} The current composition table
contains 14 rules over 18 patterns (coverage $\approx 4\%$ of all
possible pairs). The vast majority of compositions are undefined ($\bot$).
Completeness could be improved by inferring additional rules from
historical event co-occurrence data.

\textbf{Semigroup vs.\ group.} The current structure is a partial
semigroup, not a group (not all elements have defined inverses in the
composition table, and the \texttt{inverse\_table} covers only bilateral
pairs). Extending to a full algebraic group structure would allow more
powerful invariant analysis.

\textbf{Empirical validation.} The presented results are structural
predictions from the composition algebra, not forecasts validated
against historical outcomes. A rigorous evaluation would require
backtesting against documented geopolitical trajectories with
ground-truth resolution dates.

\section{Conclusion}

We have presented EL-DRUIN, a system that grounds geopolitical
relationship trajectory forecasting in formal ontology and algebraic
structures. The central contribution is the \textbf{finite semigroup forward
simulation}: starting from a specified initial pattern set, the system
iterates a composition operation defined by an explicit, algebraically
validated table, converging to idempotent attractor states that
constitute the long-run prediction. The \textbf{Lie algebra embedding}
provides a continuous-valued plausibility weight ($\mathit{lie\_similarity}$)
for each discrete composition step, coupling the discrete algebraic
structure with a continuous geometric one. \textbf{Bayesian step decay}
provides calibrated, monotonically decreasing confidence that reflects
accumulating temporal uncertainty without relying on LLM self-report.
The system is fully open-source, with a working frontend that exposes
all computation traces. We believe this architecture---algebraic
structure as the primary reasoning engine, language model as a
constrained interpreter---represents a promising direction for the next
generation of AI-assisted geopolitical analysis systems.

\bibliographystyle{plain}
\bibliography{main}

\end{document}